\def\year{2019}\relax
\documentclass[letterpaper]{article} 
\pdfoutput=1
\usepackage{aaai19}  
\usepackage{times}  
\usepackage{helvet}  
\usepackage{courier}  
\usepackage{url}  
\usepackage{graphicx,epstopdf}  
\usepackage{booktabs} 
\usepackage{multirow}
\usepackage{subfigure}
\usepackage{mathrsfs}
\usepackage[ruled,vlined]{algorithm2e}
\usepackage{algorithm2e}
\usepackage{graphicx}
\usepackage{balance}
\usepackage[labelfont=bf]{caption}
\usepackage{floatrow}
\usepackage{amsmath}
\newtheorem{mdefinition}{Definition}
\newtheorem{problem}{Problem}

\begin{document}
\title{DeepDPM: Dynamic Population Mapping via Deep Neural Network}
\author{
Zefang Zong\thanks{Equal contribution.}, \  Jie Feng\textsuperscript{$\ast$},  \  Kechun Liu, \  Hongzhi Shi, \ Yong Li\thanks{Corresponding author.}\\
Beijing National Research Center for Information Science and Technology\\
Department of Electronic Engineering, Tsinghua University, Beijing, China\\
\{zzf15,feng-j16,lkc15,shz17\}@mails.tsinghua.edu.cn, liyong07@tsinghua.edu.cn\\}

\maketitle
\begin{abstract}
Dynamic high resolution data on human population distribution is of great importance for a wide spectrum of activities and real-life applications, but is too difficult and expensive to obtain directly. Therefore, generating fine-scaled population distributions from coarse population data is of great significance. However, there are three major challenges: 1) the complexity in spatial relations between high and low resolution population; 2) the dependence of population distributions on other external information; 3) the difficulty in retrieving temporal distribution patterns. In this paper, we first propose the idea to generate dynamic population distributions in full-time series, then we design dynamic population mapping via deep neural network(DeepDPM), a model that describes both spatial and temporal patterns using coarse data and point of interest information. In DeepDPM, we utilize super-resolution convolutional neural network(SRCNN) based model to directly map coarse data into higher resolution data, and a time-embedded long short-term memory model to effectively capture the periodicity nature to smooth the finer-scaled results from the previous static SRCNN model. We perform extensive experiments on a real-life mobile dataset collected from Shanghai. Our results demonstrate that DeepDPM outperforms previous state-of-the-art methods and a suite of frequent data-mining approaches. Moreover, DeepDPM breaks through the limitation from previous works in time dimension so that dynamic predictions in all-day time slots can be obtained.
\end{abstract}

\section{Introduction}

Obtaining high-resolution population distribution(HRPD) is of great importance for urban applications of business locating, transportation planning, city service managing, etc. However, the HRPD is only available to some specific companies like the internet service providers. The general businesses have to pay a lot for it or rely on some coarse and static population data to make decisions. Meanwhile, the real-time collection of HRPD data is also computing-consuming and generates unnecessary burdens to computing systems. Thus, more efficient methods are highly required in obtaining HRPD data. 

 
 Some previous studies utilized external information like remote-sensing data to generate HRPD \cite{Wu2005Population,Gaughan2013HighRP,stevens2015disaggregating}. As the state-of-the-art for these works, R. Stevens. et.al.~\cite{stevens2015disaggregating} used random forest algorithm via lots of data sources. However, these works heavily rely on many ancillary data sets that are expensive to obtain, difficult to process, and limited to certain time slots. 

Other studies developed interpolation algorithms for super-resolution images, which provided powerful tools to convert low-resolution data into high-resolution data, independent of other ancillary data \cite{Nasrollahi2014Survey,Yang2014SingleImageSA,Vandal2017DeepSD}. One of the state-of-the-art method is SRCNN~\cite{Dong2016ImageSU}, which first utilized convolutional neural network as the interpolate function to model complex spatial relations in image. However, the amplification of this model is limited to $2\sim4$, which is too small for our population application. Additionally, the HRPD is not only related to the coarse population distribution but also the structure of the urban, which can not be considered directly by these existing methods.

Several challenges exist in dynamic population mapping. First, it is not easy to define a simple math function to describe the complex spatial relations. Second, the HRPD is influenced by other external knowledge like urban structures. Third, generating temporal trends from scratch is difficult.

In this paper, we propose DeepDPM, a deep learning based model that consists of augmented stacked super-resolution convolution neural network(SRCNN) as the static part, and time-embedded LSTM as the dynamic part, to evaluate population mapping in both spatial and temporal dimensions. Our static part predicts HRPD at different time slots, while the static output is further used by the dynamic part to generate temporal trends. PoI(Point of Interest) is also considered as an ancillary data which is easy to obtain. Our experiments showed that DeepDPM outperforms other traditional methods, and generates high-resolution results for urban population structures.

Our contributions can be summarized as follows:
\begin{itemize}
    \item We present the idea to generate, in a scalable manner, dynamic population distributions in full time series from static disaggregate data sets into finer scales. To the best of our knowledge, this is the first time to analyze both spatial and temporal patterns in an urban population mapping research work.
    
    \item We propose DeepDPM, an augmented structure that consists of static prediction and dynamic generation parts, using super-resolution convolutional neural network and time-embedded LSTM separately, based on observational and augmented PoI data.
    
    \item We perform extensive experiments based on real-life mobility dataset in Shanghai. Our results demonstrate that DeepDPM outperforms the previous state-of-the-art models and a suite of frequent data-mining and machine learning methods in terms of several metrics in predictive performance.

\end{itemize}


\section{Preliminaries}
In this section, we review the population mapping problem and introduce the special scenarios to be investigated in this paper. Then we briefly overview our solutions.

\subsection{Definitions and Problem Formulation}

\begin{mdefinition}[Grid Region]
In this study, we partition a city into an $M{\times}N$ grid map based on the longitude and latitude, where each grid denotes a region called grid region. As the basic space unit, we investigate the population distributions among these gird regions and a typical grid region is a $1km{\times}1km$ grid in the map.
\end{mdefinition}

\begin{mdefinition}[Population Distribution]
In our paper, the population distribution $X_{i}$ represented in terms of grid map depends on different aggregation level, where $i\in\{1,2,3\}$. $i$ can be equal to 1,2 or 3 each in district level, street-block level and fine-grained level, where $X_{3}$ can also be denoted as $X_{fg}$. For any pair of $X_{p}$ and $X_{q}$, $p<q$, it obeys the following relationship:

\begin{gather}
    \sum\limits_{i=x_{start}}^{x_{end}}\sum\limits_{j=y_{start}}^{y_{end}}X_{p}(i, j) = \sum\limits_{i=x_{start}}^{x_{end}}\sum\limits_{j=y_{start}}^{y_{end}}X_{q}(i, j)
\end{gather}

Further more, taking the regularity and mobility of population into consideration, we also utilize typical population trend along time to generate dynamic population distribution $X_{i}^t$ of time $t$.
\end{mdefinition}

\begin{problem}
Given the static aggregated population data $X_i$ and the regularity pattern of population, generate the dynamic finer-scaled grid region population distribution $X_{j}={X_{j}^{1},X_{j}^{2},...X_{j}^{24}}$ by a certain mapping function $F$ in a day.
\end{problem}

\begin{figure}[ht]
    \centering
    \includegraphics[width=0.95\columnwidth]{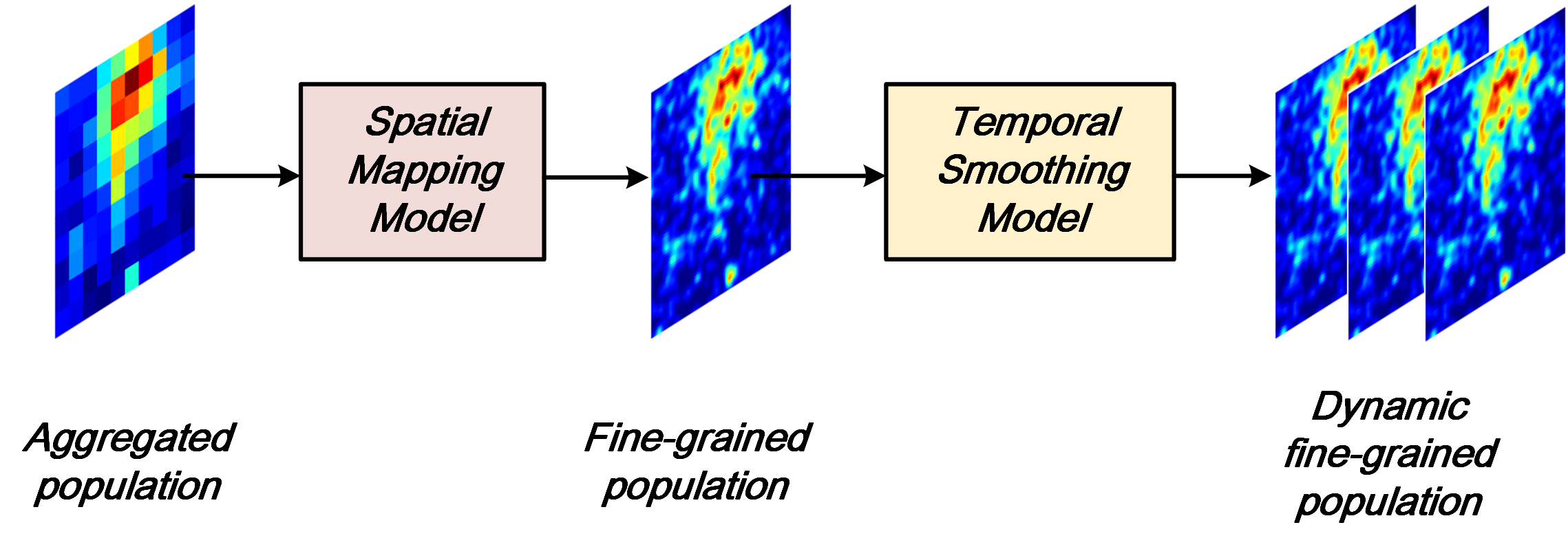}
    \caption{Basic framework of our solution for aggregated population mapping problem.}
    \label{fig:intuition}
\end{figure}

\subsection{Solution Overview}
As Figure~\ref{fig:intuition} shows, we divide the total task into two sub-tasks as follows.

\textbf{Problem 1: Spatial modelling.} Previous studies are limited in the reality. Existing studies~\cite{Gaughan2013HighRP,stevens2015disaggregating} rely on various ancillary data like high resolution imagery to obtain the spatially weighted density. However, these large-scale data are expensive to obtain and challenging to process. Furthermore, because of the strong dependency on the spatial information from the input data, previous studies used semi-automated classification algorithms combined with simple dasymetric mapping approach like random forest~\cite{stevens2015disaggregating}, which is limited in directly modelling the spatial relations. 

\textbf{Solution Overview:}As a powerful spatial modelling tool, convolution neural network (CNN)~\cite{LeCun1989Hand,He2016DeepRL} is widely used in many tasks. Particularly, CNN has been applied into image super-resolution task, which is aimed to generate high-resolution image based on the low-resolution image. Based on the formulation in the previous section, the aggregated population mapping task follows the similar goal and working manner with image super-resolution task. Inspired by this, we introduce CNN-based model into our task to enhance the modelling of spatial relations and constrains that exist. 

\textbf{Problem 2: Temporal modelling.} Although the temporal trend is important, little previous studies of aggregated population mapping task has considered this because of the lack of dynamic data and proper methods for generating the temporal trend of population distribution. They regarded the population mapping problem as a static process only considering the night scenarios. However, the fact is that population distributions in the day in city are totally different from those in the night. Thus, solutions proposed by previous studies failed in generating the fine-grained population distribution during a full day long period.  

\textbf{Solution Overview:} According to Xu. et.al.~\cite{Xu2016ContextawareRP}, the temporal pattern of population for a certain region can be divided into several typical classes based on its function. For example, the population of residence region first decreases to a low level in the morning, keeps during the day and finally comes back to the original level. Thus, the temporal pattern of population can be regarded as a function of time. Based on the observation, we consider to utilize recurrent neural network (RNN)~\cite{Lipton2015A,Hochreiter194LSTM} as the basic sequential model to generate temporal trends for population of each region. Particularly, we introduce the time factor into the model by embedding to control the generation process.

\section{The Spatial-temporal Mapping Model}
\begin{figure}[ht]
    \centering
    \includegraphics[width=0.95\textwidth]{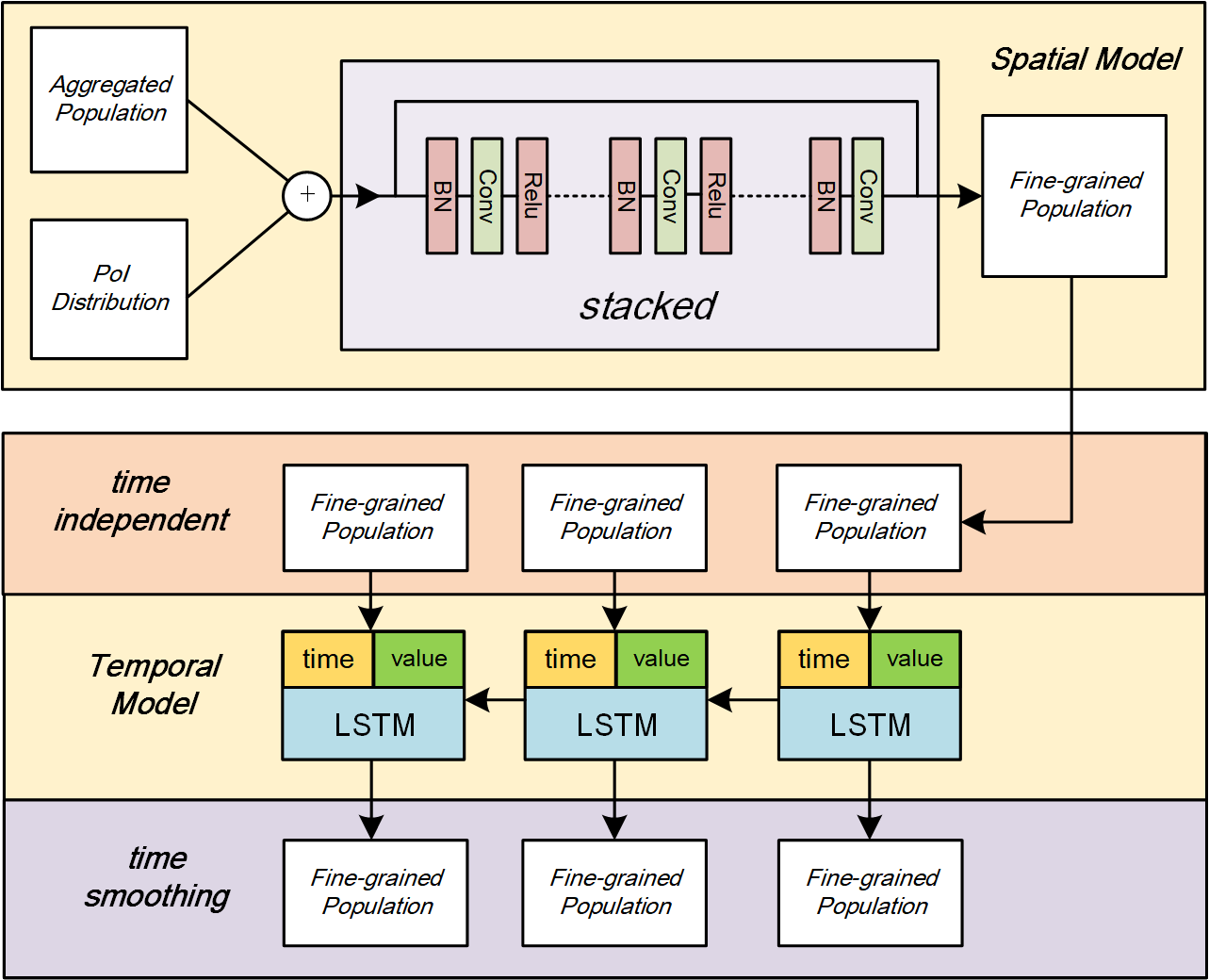}
    \caption{The Network structure of our spatial-temporal mapping model.}
    \label{fig:model}
\end{figure}

 As presented in the Figure~\ref{fig:model}, our model consists of two main components: 1) spatial mapping model, which is designed to map the aggregated low-resolution population into high-resolution population; 2) temporal generation model, which is designed to capture the typical temporal trend and generate smoothed dynamic population results. 

\subsection{Spatial Mapping Model}
Mapping the aggregated population into higher-resolution population is similar to image super resolution problem, in which SRCNN is one of the state-of-the-art methods.

SRCNN consists of three operations ~\cite{Dong2016ImageSU}: patch extraction, non-linear mapping, and reconstruction. In SRCNN, each operation is implemented as a three-layer convolution network including a batch-norm layer, a convolution layer and a non-linear activation layer (e.g., Relu). By stacking these three operation units, we formulate basic spatial mapping unit in our model. With the low-resolution image $X$ as the input and the high-resolution image $Y$ as the target, the mapping function $F$ is optimized with the following objective function:
\begin{gather}
    \arg\mathop{\min}_{\Theta}\sum^n_1{{\lVert F(X_i;\Theta) -Y_i \rVert}_2^2},
\end{gather}
where $\Theta$ denotes the parameters of the network and $n$ denotes the instances of image pairs.

\begin{figure}[ht]
    \centering
    \includegraphics[width=0.95\textwidth]{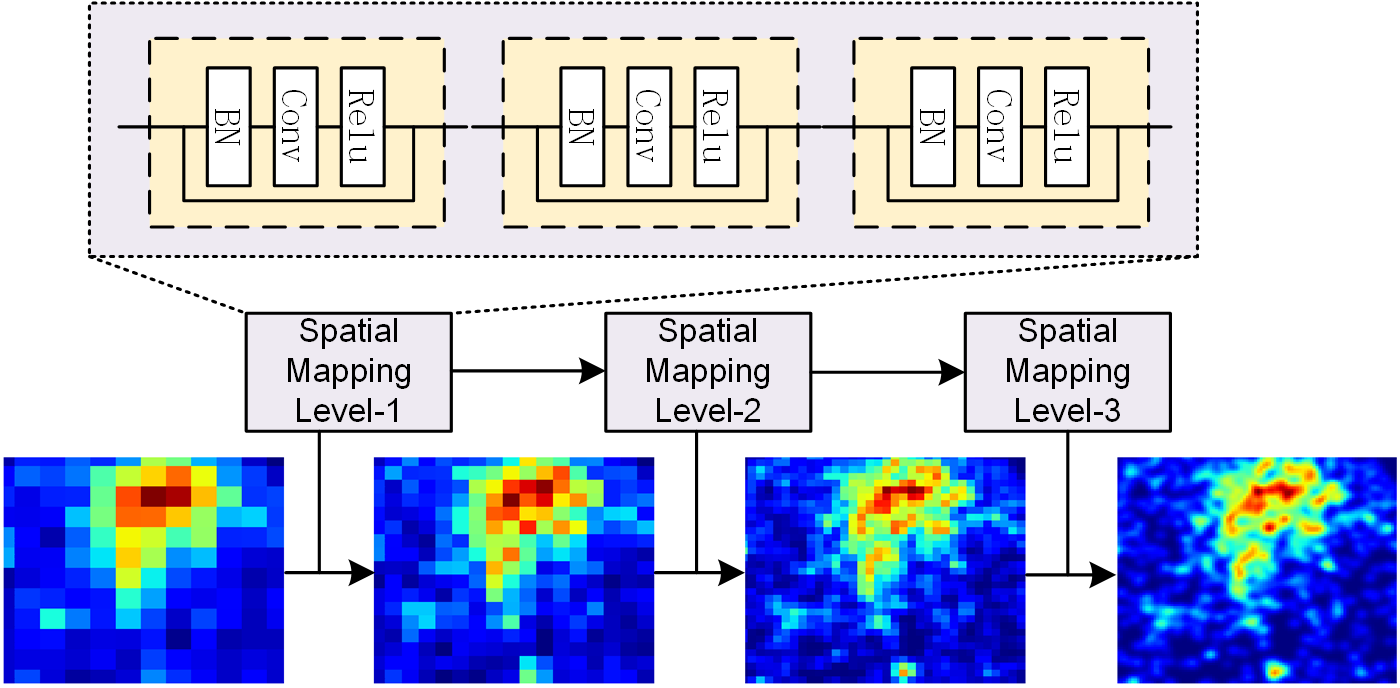}
    \caption{Stacked network structure of the spatial mapping model.}
    \label{fig:spatial-model}
\end{figure}

The basic mapping unit SRCNN can only handle the resolution enhancement ratio between 2 to 4, while in population mapping this ratio can be up to 15. One possible solution is to stack more convolution networks to directly meet with the higher enhancement requirement, which however, fails in learning complex mapping functions because of the lengthy propagation path and weak supervised signals. Thus, we decompose the high-enhancement ratio mapping task into several low-enhancement ratio tasks. In practice, we train several independent mapping network units for each of these sub-tasks by providing their related mapping ground-truth. Finally, we stack these trained independent mapping units to form a comprehensive spatial mapping model to complete the whole mapping task. In detail, two SRCNNs are stacked in generating $X_2$ from $X_1$, and another two are stacked in generating $X_3$ from $X_2$. The basic mapping units and the whole procedure are presented in Figure~\ref{fig:spatial-model}.

It's worthy to mention that cascading super-resolution networks altogether directly is also a considerable method, which is to train the model in one end-to-end way. However, during experiments we found that stacking outperforms cascading. Generating distributions in a lower resolution by upscaling fine-grained ground truth allows us to train independent input/output pairs and stack them together at test time as well as keeping accuracy. While in a cascading one, the fact that each output is exactly the input of the following level may lead to some error in propagation\cite{Vandal2017DeepSD}.
Besides, according to Xu. et.al.~\cite{Xu2016ContextawareRP}, the function of a region can play an import role in forming its population pattern. Meanwhile, the function of a region can be represented by the Point of Interests (PoIs) distribution to some extent. Hence, we introduce PoI distribution matrix to describe the function of regions. Different types of PoI matrix are regarded as specific channels to form the multi-channel input matrix in aggregated level.

\begin{figure}[ht]
    \centering
    \includegraphics[width=0.95\textwidth]{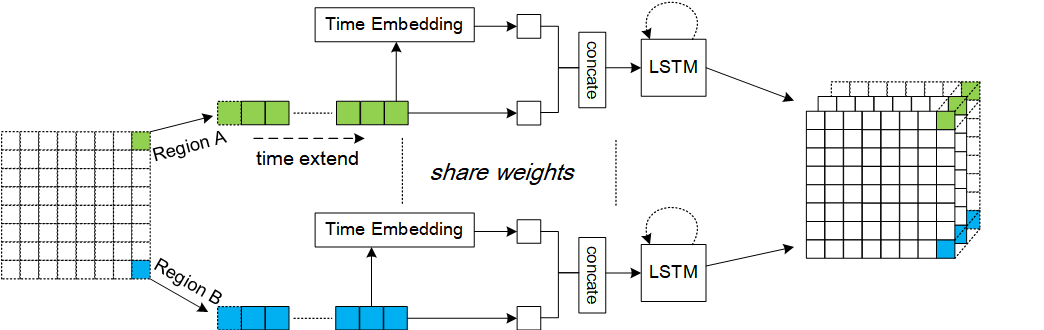}
    \caption{Structure of temporal smoothing model.}
    \label{fig:lstm}
\end{figure}


\subsection{Temporal Generation Model}
Except the spatial distribution, the temporal trend is also an import characteristic of population distribution. By smoothing population along time, we can reduce the spatial estimation error and better understand the distribution. According to Xu. et.al.~\cite{Xu2016ContextawareRP}, while the concrete temporal variations of population in different regions are different, their temporal trends can be classified into several typical types based on the function of each region. And these temporal trends are representative and shared by different regions and cities. In this paper, we use long-short term memory network (LSTM), a population variation of RNN as the basic recurrent unit to model these typical temporal trend series. 


During the training, we first extract the temporal population series of every region from the output of spatial mapping model. Then, we use one LSTM model to train all the population series. In this way, the parameters of the neural network are shared by all regions, which makes the model small, robust and can be generalized in modeling the temporal trends of population. To model the influence of time factor (e.g., hour of day), we introduce time embedding into our temporal generation model. Particularly, we first discrete the time of day into 24 hours and encode them into 24-dimension one-hot vectors. Then, we build a linear network layer as embedding table to project the one-hot vector into dense vector. 
Then, the original population value (unknown value can be set as 0) and the time vector are grouped together as a whole to feed into LSTM network in every time step. The details of the temporal structure is presented in Figure~\ref{fig:lstm}.




\section{Performance Evaluation}

In this section, we conduct extensive experiments on mobility dataset in Shanghai to answer the following research questions:

\begin{itemize}

    \item \textbf{RQ1: Spatial modelling to obtain the population distribution in higher resolutions at a fixed time slot} 

    \item \textbf{RQ2: Temporal modelling to obtain the dynamic population distribution in a fixed resolution}

\end{itemize}

\subsection{Experimental Settings}

\subsubsection{Datasets}

We collect our representative real-life mobility dataset from ISP , which contains cellular network access records in 9685 different base stations in Shanghai for 4464 different time slots, from 1$^{st}$ July, 2017 to 31$^{st}$ July, 2017 (the data usage is recorded every 10 minutes for 31 days). Considering the periodicity difference between weekdays and weekends, we manually drop the data on weekends and focus on weekdays' data in experiments. 

We use the address of base station to estimate cellular data users' location distribution. Since mobile devices keep accessing cellular data as long as their data connection are kept on, our dataset well represents the population distribution in Shanghai. Similar data types are used in urban research as well, such as call detail records \cite{isaacman2012human,ficek2012inter} or GPS data\cite{zheng2008understanding}. However, these data above are event driven, which update only when a user acquires service. While our dataset passively captures users' newest location information, which guarantees the credibility of our analysis.

Also, we collect our PoI dataset from Tencent, which contains 618296 PoI records in 17 categories. We manually classify them into 4 categories, entertainment, business, transportation junctions and residence, based on their functions.

\begin{table}[h]
    \label{tab:poi classification}
    \resizebox{3.3in}{!}{
    \begin{tabular}{cl}
        \hline
        \textbf{Aggregated Categories} & \multicolumn{1}{c}{\textbf{ Original Categories}}
        \\\hline
        Entertainment       & \begin{tabular}[c]{@{}l@{}}
        \small{Hotel, Entertainment, Shopping,}\\
        \small{Catering, Culture, Sports,}\\
        \small{Tourist Spots}\end{tabular}
        \\\hline
        Business and Education      & \begin{tabular}[c]{@{}l@{}}
        \small{Departments, Industries,}\\
        \small{Education, Medical}\end{tabular}
        \\\hline
        Transportation Junctions       & \begin{tabular}[c]{@{}l@{}}
        \small{Transportation Junctions}\end{tabular}
        \\\hline
        Residence       & \begin{tabular}[c]{@{}l@{}}
        \small{Housing, Residential Services}\end{tabular}
        \\\hline
    \end{tabular}}
    \caption{PoI Classification from original categories into aggregated categories.}
\end{table}


\subsubsection{Preprossessing}

One obstacle in using mobility records to represent population distribution is that base stations lose the track when devices are turned off, or are disconnected due to other various factors, like weak signal strength. Therefore, an augmented algorithm is in need to recover the missing fingerprints. 

We define the record user number of base station $i$ in time slot $t$ as $X_{i}[t]$, the actual number as $Y_{i}[t]$, the total time slot number as $T$, and the total station number as $N$. We first compute the sum of all activated devices at each time slot $t$, and find out the maximum of it ever recorded as the representation of population amount in the city. Then we estimate the percentage of activated devices denoted by $R_{i}[t]$. Finally we obtain the estimated user number $Y_{i}[t]$. The formulation in math is described below:

\begin{gather}
S[t] = \sum\limits_{i=1}^{N}X_{i}[t]\: for\: t = 1\:to\: T,\\
R[t] = \frac{S[t]}{\max_{j=1}^{T}S[j]}\: for\: t = 1 \:to \:T,\\
Y[t] = \frac{X_{i}[t]}{R[t]},\: for\: t = 1\:to \:T, \:i = 1\: to\: N
\end{gather}

Since the base stations are located irregularly in geometry, we need to further generate the grid regions. We generate Voronoi diagrams based on the distribution of base stations. The contribution of population from each polygon to each grid is determined by the ratio, which is the intersection area divided by the polygon area. Finally, the population distribution at each time slot can be successfully mapped into grids, which is an 83*114 grid map based on longitude and latitude. PoIs mapped into the same grid are counted together, and the sum is assigned as the grid value according to different PoI categories.

After obtaining the fine-grained grid region distribution denoted as $X_{fg}$ or $X_3$, we further generate the aggregated distribution at another two levels, $X_{1}$ and X$_{2}$. The grid maps remain the same size, while grids in the same district or the same street area are equalized using their average. Grid values outside the boundary are set to 0. Those grids where more than half of the area are out of boundary are dropped from the patch set.

\subsubsection{Baselines}
Automated Statistical Downscaling(ASD) \cite{hessami2008automated} is a traditional method for statistical downscaling. ASD requires regression methods to predict population density pixel by pixel. We compared three ASD methods, which are logistic and lasso regression, support vector machine(SVM) regression and artificial neural network(ANN) regression. Each method uses the density of lower resolution and PoIs to predict the higher resolution population map. Due to the time complexity of SVM, we randomly chose 80000 pixels to train SVM model.
A second set of methods, random forest-based dasymetric mapping approach~\cite{stevens2015disaggregating} and decision tree algorithm are applied to compare to our spatial mapping model. According to the approach described by Stevens et al.~\cite{stevens2015disaggregating}, all the population data are transformed into log density. Higher resolution map is predicted by applying random forest regression on log population density of lower resolution and PoIs. Decision tree algorithm uses the same data processing approach.

\subsubsection{Metrics and Parameter Settings}
We use 5-fold cross-validation in the experiment. For static model that consists of SRCNNs, the input data is obtained by concatenating population-grid maps with PoI-grid maps. The depth of the final matrix relies on how many PoIs categories we use, with all 4 categories as default. Except that 38$\times$38 patches are used in $X_2$-$X_{fg}$ level and 58$\times$58 in $X_1$-$X_2$ level, all SRCNNs are trained with the same set of parameters.Layer 1 consists of 64 filters of 9x9 kernels, layer 2 consists of 32 filters of 1x1 filters, and the output layer uses a 5x5 kernel. Higher resolution models which have a greater number of sub-images may gain from larger kernel sizes and an increased number of filters. Each network is trained using Adam optimization with a learning rate of $10^{-4}$ for the first two layers and $10^{-5}$ for the last layers, and MSE loss as the loss function for every training step.

Each model is trained for 10${^5}$ iterations with a batch size of 512. Tensorflow is utilized to build and train DeepDPM. Each SRCNN is trained independently on a Titan X GPU, and the inference is then executed sequentially on a single Titan X GPU.

In order to measure the performance of our structure and other traditional methods in comparison, we use a few key metrics to show static model's applicability. Root mean square error(RMSE) and Pearson's correlation(CORR) are used to measure the prediction quality. We also use normalized root mean square error(NRMSE) for inner comparison later. 


\subsection{RQ1: static population mapping performance}
\subsubsection{Overall Model}
Without considering dynamic changes in distribution during the day, we first train an overall model with all data available in weekdays to compare DeepDPM with other baseline methods. 

Our experiment compares performance with another six approaches, static model, random forest, decision tree, svm, ann and lasso, presented on Table~\ref{tab:overall performance} The three metrics discussed above are computed at all time slots in the test set and the averages are collected. We find that our static model outperforms all other methods in both $X_1$-$X_2$ level and $X_2$-$X_3$ level in terms of all three metrics. In detail, random forest gives the best prediction among all traditional methods, and is slightly outperformed by our static model by a difference in RMSE for no more than 3.0 in $X_1$-$X_2$ level. While the difference enlarges as the number of stacked SRCNNs increases, which is about 14.6 in $X_1$-$X_3$ level. In terms of correlation, decision tree, random forest and our model all perform well. Ann costs the longest run time, while it performs poorly compared to others. Stacked convolution neural network shows its strong ability in describing spatial structure.

\begin{table*}[h]
\small
\centering
\begin{tabular}{ccccccc}
\toprule
& \multicolumn{3}{c}{District to Street-Block($X_1$-$X_2$)} & 
\multicolumn{3}{c}{District to Fine-Grained($X_1$-$X_3$)}\\
Method&RMSE&NRMSE&Corr&RMSE&NRMSE&Corr\\
\midrule
Lasso&513.2714&2.5975&0.7966&697.3197&3.5237&0.7144\\
ANN&465.7865&2.3573&0.8362&679.8032&3.4352&0.7334\\
SVM &850.5768&4.3045&0.2658&1002.3468&5.0650&0.2215\\
DecisionTree &51.5804&0.2610&0.9982 &117.1829&0.5921&0.9931\\
Random Forest &47.0408 &0.2381 &0.9985&93.5023 &0.4725 &0.9956 \\
Static Model&\textbf{44.5574}&\textbf{0.2255}&\textbf{0.9987}
&\textbf{78.9081}&\textbf{0.3987}&\textbf{0.9978}\\
\bottomrule
\end{tabular}
\caption{Comparison of predictive ability between all six methods for all time slots in the dataset. All four PoIs are used in the experiments.} \label{tab:overall performance}

\end{table*}

\subsubsection{Poi Influence on Model Performance}
\begin{figure}[h]
    \centering
    \subfigure[RMSE performance on different poi combinations.]{
        \includegraphics[width=0.95\textwidth]{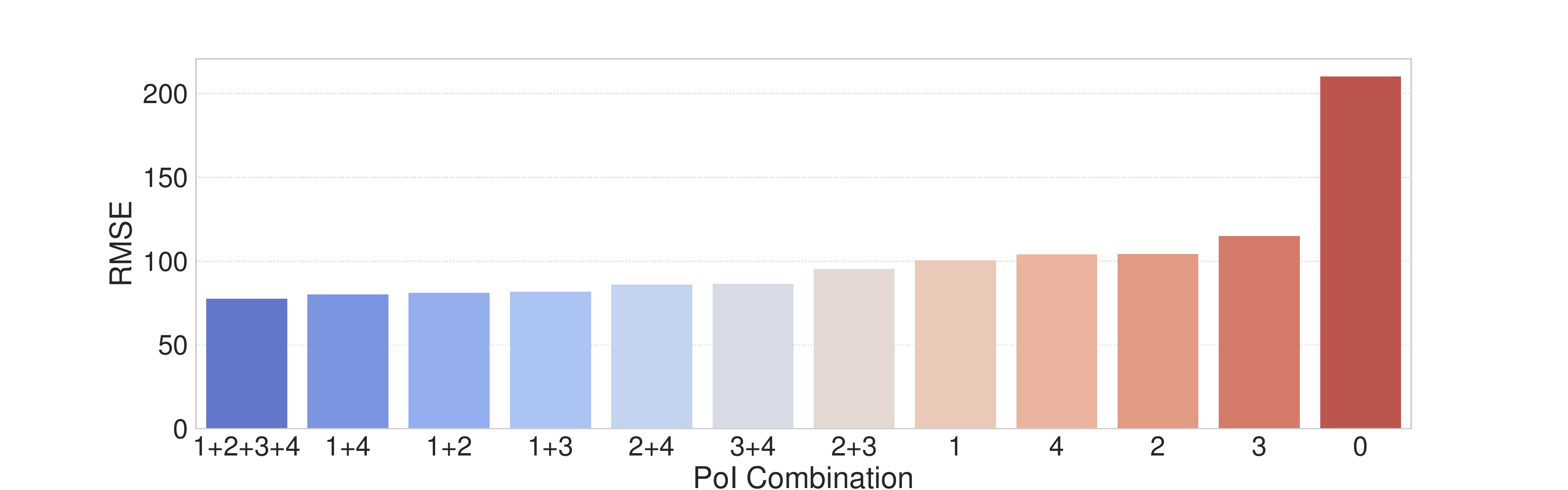}
        \label{fig:poi_combination2}
    }
    \subfigure[Results on different poi distribution.]{
        \includegraphics[width=0.95\textwidth]{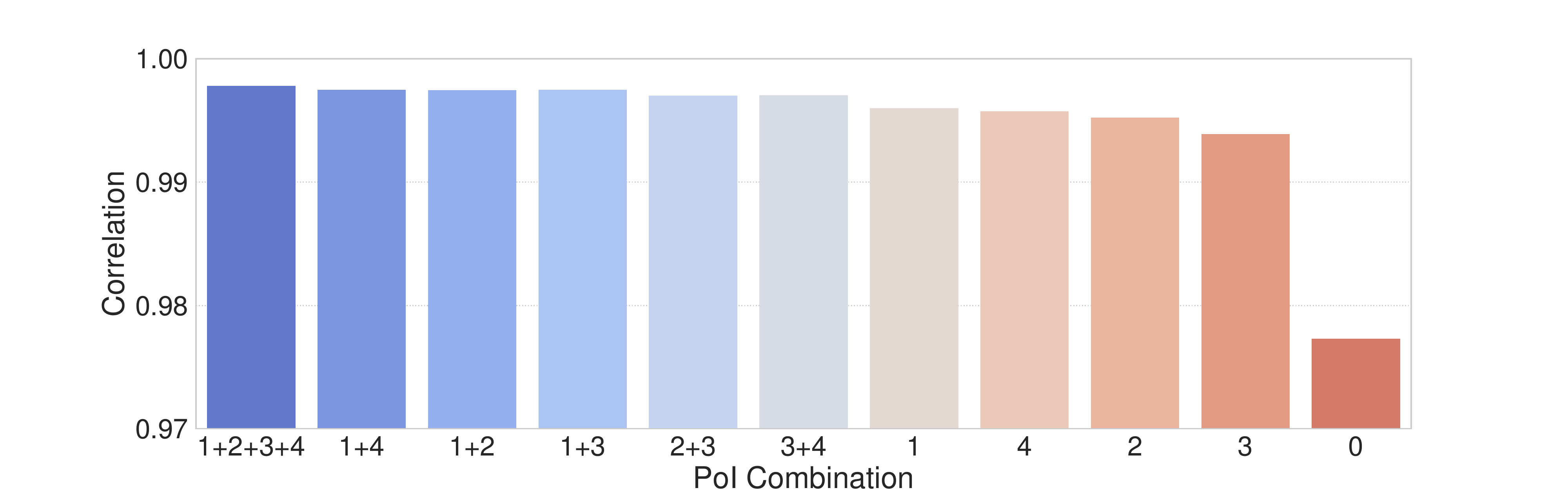}
        \label{fig:poi_combination1}
    }

    \caption{Comparison of predictive ability using DeepDPM with differnet PoI combinations. \#1, \#2, \#3, \#4 stand for entertainment, business, transportation, residence PoIs separately. \#0 stands for prediction with completely no PoI usage.}
    \label{fig:poi_combination}
\end{figure}

It is important to choose correct type and amount of PoIs as augmentation before training. We run our experiment based on different combinations of PoI and with completely no PoI as presented in Figure~\ref{fig:poi_combination} 
Generally, performance gets promoted rapidly as PoI usage increases. The result verifies the hypothesis that the more information augmented in PoIs we add into our model, the more precise our predictions will be. Using all four categories gives the best result. In detail, we find out that the entertainment PoIs play the most important role, with residential PoIs following when different categories are considered alone. The combination usage of such two PoIs also prove to outperform other bi-combinations. The model without using any PoI performs terribly. Local functions of different regions prove to be an important factor in describing population distribution pattern.


\begin{figure}[h]
    \centering
    \subfigure[Results on downtown and suburb.]{
        \includegraphics[width=0.3\textwidth]{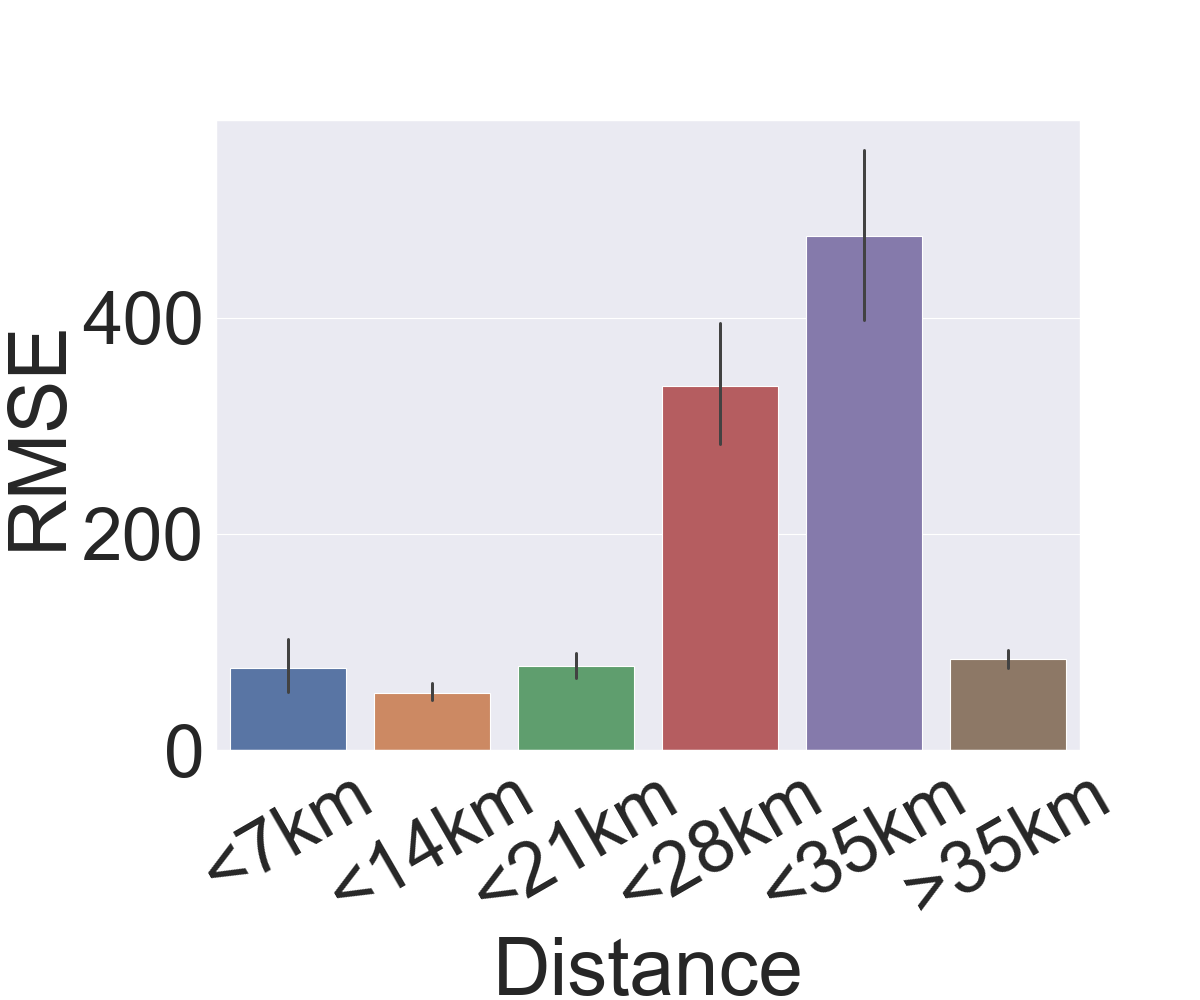}
        \label{fig:distance}
    }
    \subfigure[Results on different PoI distribution.]{
        \includegraphics[width=0.3\textwidth]{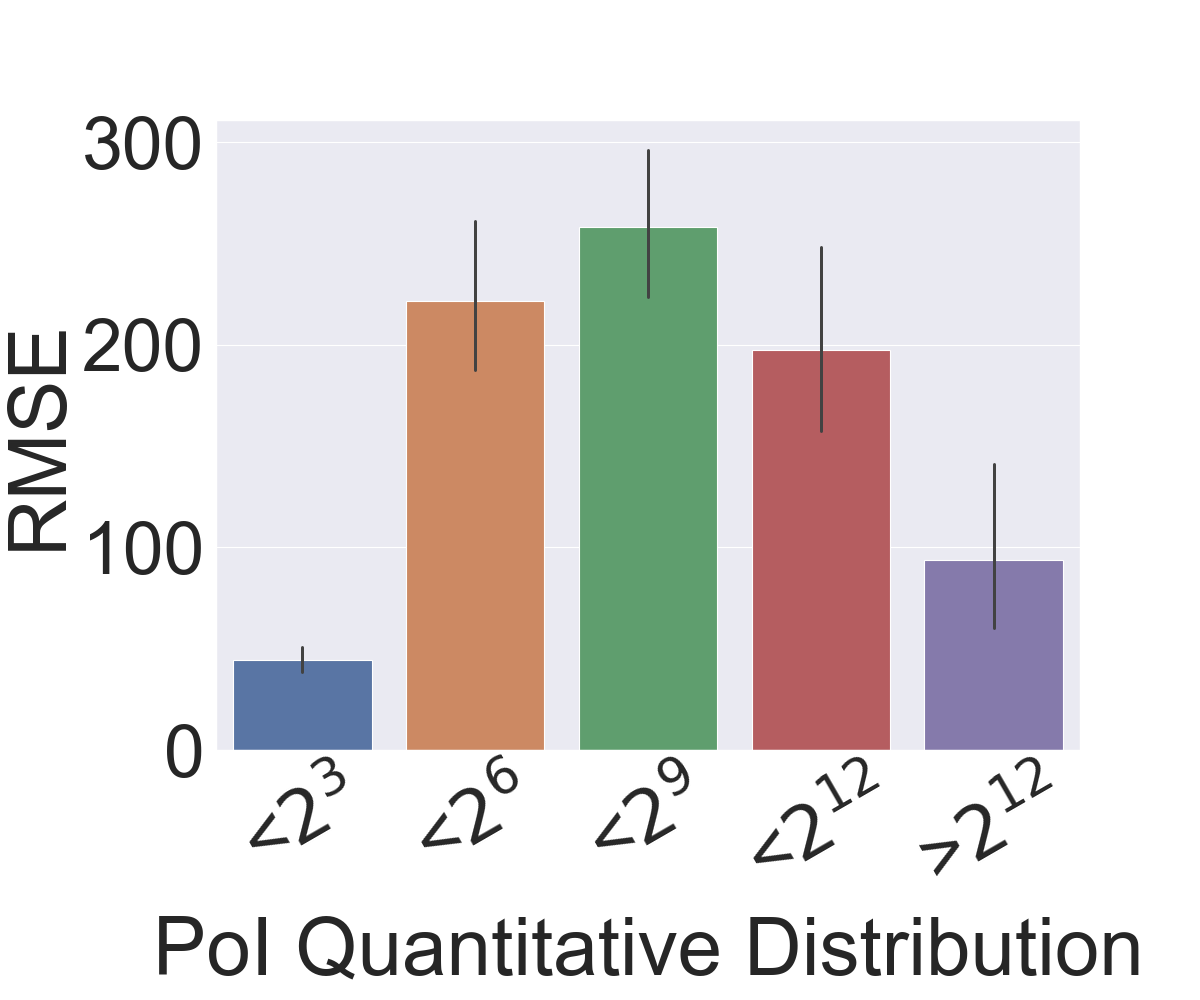}
        \label{fig:poi}
    }
    \subfigure[Results on different functional zones.]{
        \includegraphics[width=0.3\textwidth]{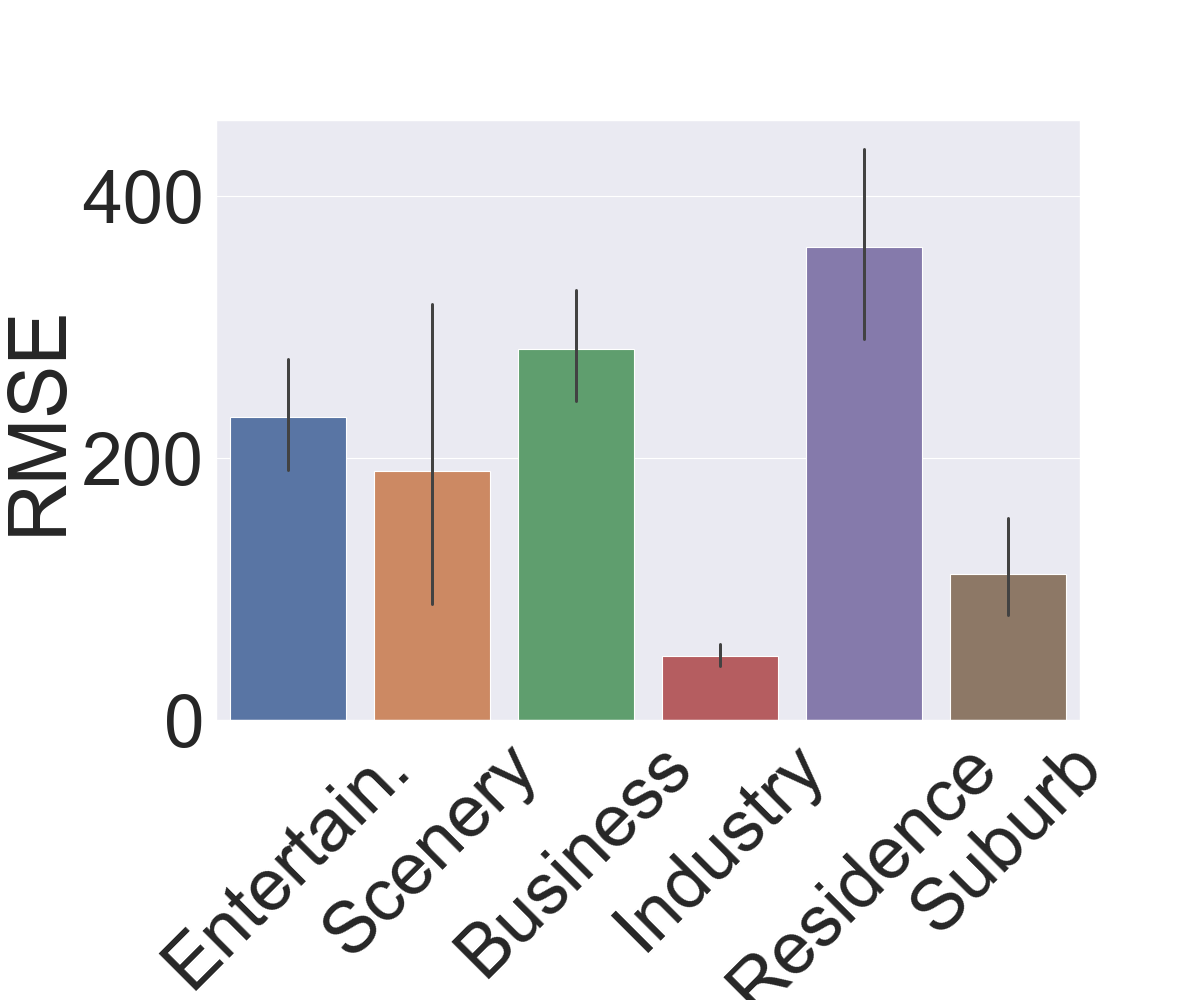}
        \label{fig:function}
    }
    \caption{Comparison of local prediction performance using DeepDPM in different regions.}
    \label{fig:variations}
\end{figure}

Besides measuring global predictive ability based on different PoI usage, we also test local performance in different regions in our default model considering all 4 PoIs, shown in Figure~\ref{fig:variations}. Figure~\ref{fig:distance} shows the relationship between the RMSE with the distance to the center of downtown (the grid where there is the highest population density). It turns out that we can reach high mapping accuracy in both suburbs and downtown areas , however the performance descends rapidly in the joint places. This is for population distribution change frequently in these places, which makes it hard to predict population distribution in finer scales. Figure~\ref{fig:poi} shows the relationship between the performance and the amount of PoIs locally. Figure~\ref{fig:function} shows the relationship between local performances with the functions of local region. Industrial parks and suburbs turns out to have a much better performance than other regions, for the population distributions in these areas are much steadier than those in other regions as time changes, while residence regions suffer from frequent population movement. 

\subsubsection{Segmented Model}

Considering different distribution patterns at different time slots in a day, temporal changes might have a strong impact on the prediction. We manually separate the entire dataset into three parts, which represent a specific period each, to further investigate the influence of different time in a day on our model precision. Period intervals include 0:00-7:00, 7:00-17:00 and 17:00-24:00. Considering the length of this paper, we only show the comparison results in Period2(7:00-17:00) in Table \ref{tab:period2 performance}.

We find that DeepDPM still outperforms other baseline algorithms in three segmented models. While compared to the overall model using all time slots, the segmented one reached better performance. It is because population is more steady within a limited period. It proves that a flatter temporal trend in a fixed period in a day helps to improve predictive ability in the static model, for the static model itself doesn't take temporal changes into account. 

However, the improvement from time slot segmentation is not enough to evaluate temporal changes in population distribution. We further put forward our dynamic population mapping model and conduct experiments to solve the problem addressed. 


\begin{table*}[htb]
\small
\begin{center}
\begin{tabular}{ccccccc}
\toprule
 & \multicolumn{3}{c}{District to Street-Block($X_1-X_2$)} & 
\multicolumn{3}{c}{District to Fine-Grained($X_1-X_3$)}\\
Method&RMSE&NRMSE&Corr&RMSE&NRMSE&Corr\\
\midrule
Lasso &491.4463&2.4870&0.8025&667.8841&3.3749&0.7214\\
ANN &441.9888&2.2368&0.8440&641.2524&3.2404&0.7762\\
SVM&824.2706 &4.1713 &0.2733&969.5835 &4.8995 &0.2285\\
DecisionTree&44.0956&0.2231&0.9986&97.1856&0.4911&0.9953\\
Random Forest&42.8038 &0.2166 &0.9986&84.4569 &0.4268 &0.9961\\
Static Model
&\textbf{40.7466}&\textbf{0.2062}&\textbf{0.9989}
&\textbf{76.9615}&\textbf{0.3788}&\textbf{0.9980}\\
\bottomrule
\end{tabular}
\caption{Comparison of predictive ability between all six methods for time slots in period 2, from 7:00 to 17:00 every day.}
\label{tab:period2 performance}
\end{center}
\end{table*}


\subsection{RQ2: dynamic population mapping performance}

\subsubsection{Quantitative Results}

We use time-embedded LSTM to generate our temporal model. Fine-grained results from static model in $X_{2}$ - $X_{3}$ level are the input sent to the model. Table~\ref{tab:SHperformance} shows the prediction performance in terms of RMSE, NRMSE and MAE. The initial results from the static model and prediction using flat LSTM are shown as baselines.

Compared with our static model, both two LSTM models showed their advantage of their powerful ability in sequence modeling. We find that the the LSTM+Time Embedding model reduces NRMSE by $14.5\%$. This suggests that the strong time-sequence regularity in population that our static model doesn't captured can be modelled well in our temporal model. As shown in (a) in Figure~\ref{fig:dynamic performance}, NRMSE changes in all three methods in all time slots of a day are illustrated. Our temporal model outperforms other two models at almost every time slot, except that LSTM performs as well as it at several moments at around 4:30 and 17:00. The general LSTM model has more accurate prediction than static mapping except from around 7:00 to 11:00. The temporal model trained based on the input from our static one can be able to model the complex sequential transition as well as holding attention on spatial patterns.
\begin{table}[h]
\small
\begin{center}
\caption{Performance comparison of static and dynamic population mapping in Shanghai.}\label{tab:SHperformance}
\begin{tabular}{cccc}
\toprule
Method&RMSE&NRMSE&MAE\\
\midrule
Static Mapping&86.87&0.4517&32.28\\
LSTM&81.46&0.4236&31.60\\
LSTM+Time Embedding&74.27&0.3862&32.01\\
\bottomrule
\end{tabular}
\end{center}
\end{table}

\subsubsection{Illustrated Cases}
\begin{figure}[ht]
 \centering
  \subfigure[Performance comparison in terms of global NRMSE at different time slots.]{
        \includegraphics[width=0.47\textwidth]{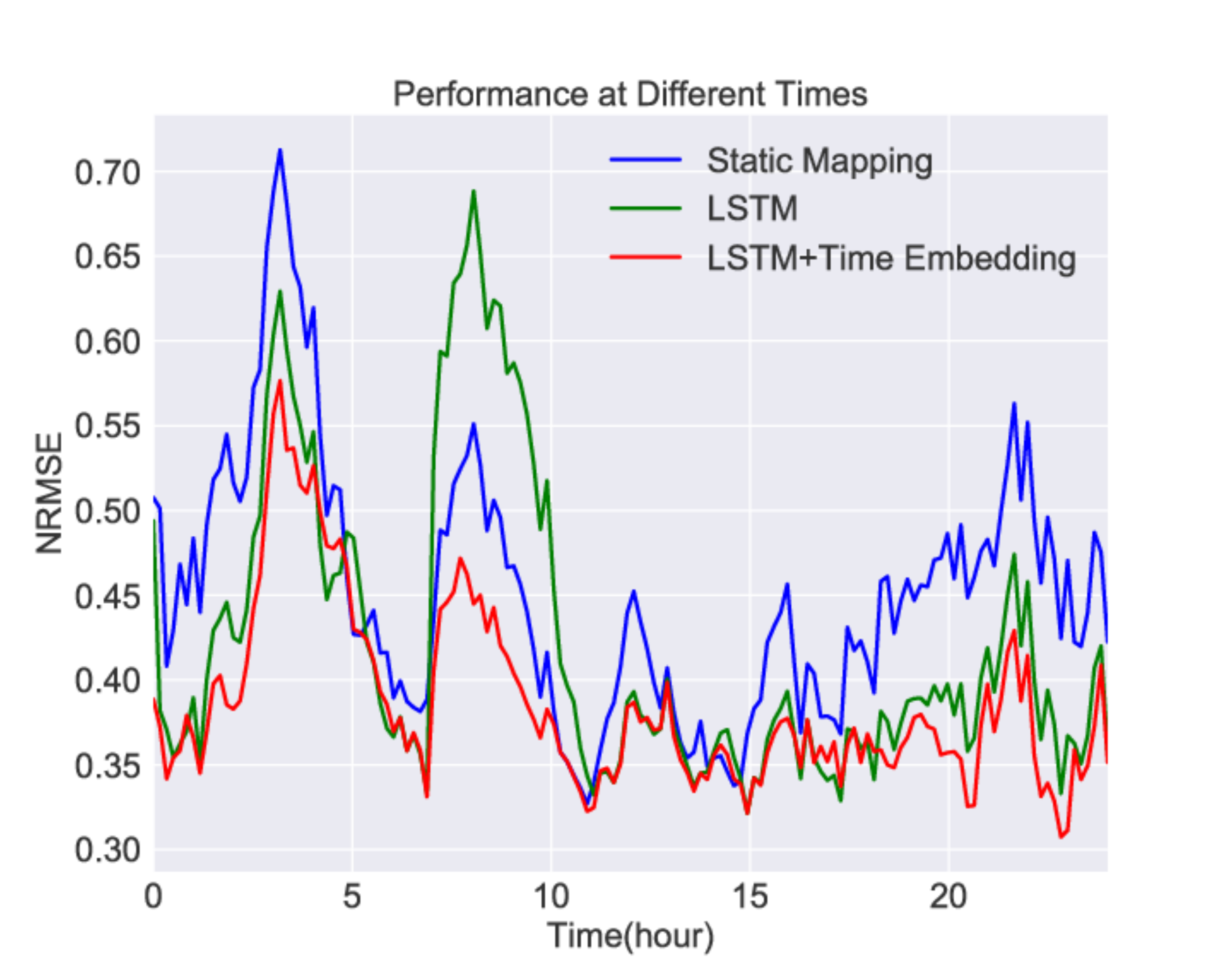}
    }
    \subfigure[Case study: population series at one typical grid for all three model predictions and ground truth.]{
        \includegraphics[width=0.47\textwidth]{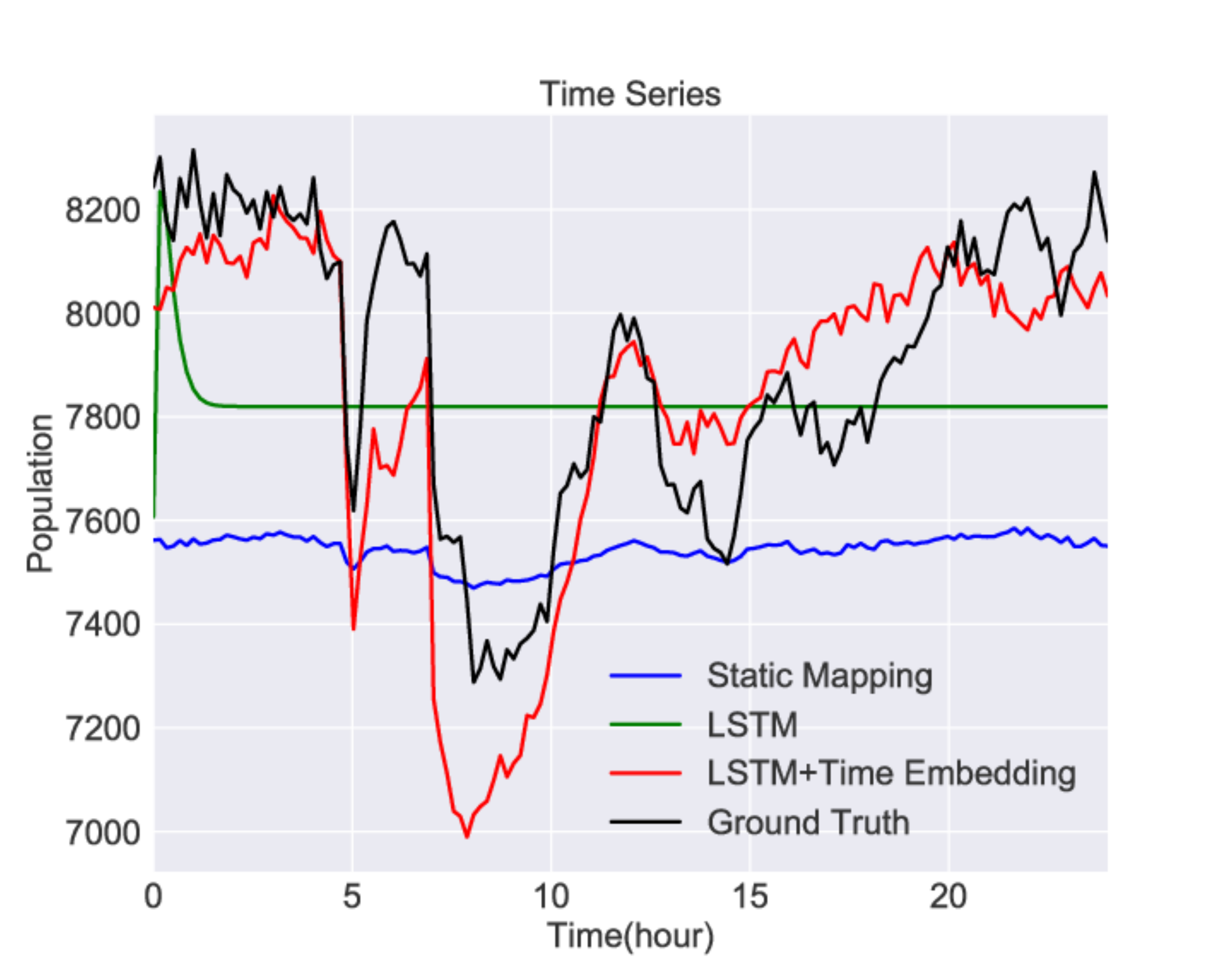}
}
 \caption{Performance of Dynamic population Mapping in Shanghai.}
 \label{fig:dynamic performance}
\end{figure}

After a generalized analysis, we focus on detailed prediction performance in local areas. We choose a typical grid to study the predicted population series in a day time. (b) in Figure~\ref{fig:dynamic performance} shows the population change in a day time. The red curve stands for the result from our temporal prediction, from which we can tell the similarity more visually. The blue curve represents prediction result from our static model, which almost remains at the same quantity, even though it shares the same undulating trend as the ground truth. The changing range in a day time of ground truth exceeds 1000 in population, while the static model only ranges no more than 100.It shows that the spatial modelling using super-resolution structure does lose sequential structures when capturing spatial regularity. While the population sequence pattern of our temporal prediction is much more similar to the ground truth.

The case study explains the reason static model fails to predict temporal trend that the temporal one is able to. Since the distribution in all time slots are regarded as the same into SRCNNs, the system tends to average all population signals from different time slots, which results in a great temporal pattern loss. While our time-embedded LSTM temporal model overcomes the shortcoming, and retrieves it by time-based training. The whole DeepDPM system thus retains both spatial and temporal patterns in urban population distribution.

\section{RELATED WORK} 
\label{sec:Relatedwork}

Two major fields are related to our study. 

\textit{Fine-Grained Population Mapping:} Early studies \cite{anderson1973population,hessami2008automated,sutton2001census} used remotely sensed information, such as satellite imagery, as the main data source. While \citeauthor{azar2010spatial,chen2002approach} chose to refine census population distribution using ancillary data. As the state-of-the-art method in the field, R.Stevens \cite{stevens2015disaggregating} used random forest \cite{liaw2002classification,breiman2001random} algorithm as a dasymetic redistribution approach based on both census and remotely sensed data. Compared to their studies, our DeepDPM uses coarse population data like census data, and PoI data as augmented data, which are much easier to obtain. Besides, our study breaks through the limitation in the time dimension.

\textit{Image Super-Resolution:} Early studies used filtering approaches, \textit{e.g.} linear, bicubic or Lanczos \cite{duchon1979lanczos} filtering. \citeauthor{freeman2002example} and \citeauthor{freeman2000learning} firstly sought to construct mapping algorithm between training patches and corresponding known high-resolution counterparts.
In recent years, convolutional neural networks(CNN) based SR algorithms have shown excellent performance \cite{wang2015deep,Dong2016ImageSU,wang2016end,kim2016deeply}, where
SRCNN \cite{Dong2016ImageSU} is one of the state-of-the-art for the problem. \citeauthor{Vandal2017DeepSD} successfully used the SRCNN based DeepSD structure in climate prediction. Our study also learns from the advantage of SRCNN to construct our static part of DeepDPM structure, and furthermore implements the dynamic part to learn the temporal pattern.

\section{LIMITATION AND FUTURE WORK}
Currently, we are using mobile dataset to represent HRPD. However, the premise of our preprocess method is the hypothesis that the urban area has no explicit inside or outside population flows. This may be a major source of systematic error. We will consider more practical approaches to quantify the gain and loss of our current method, and explore more to reduce the error.

\section{Conclusion}
\label{sec:Conclusion}
In this paper, we investigate population mapping in both static and dynamic view using PoI as augmented information. We propose a deep learning based model to generate a complete population mapping structure, DeepDPM, which has two novel characteristics compared to previous studies and methods: 1) a stacked SRCNN based static model that evaluates static population prediction; and 2) a time-embedded LSTM based dynamic model that smooths the temporal change. Extensive experiments on the dataset of mobility data collected from Shanghai showed that DeepDPM significantly improves the performance compared to all other baselines. Meanwhile, our structure also breaks through the limitation in time dimension that previous studies had. As a result, population distribution in full time series is generated.

\section{Acknowledgments} \label{acknowledg}
This work was supported in part by The National Key Research and Development Program of China under grant 2017YFE0112300, the National Nature Science Foundation of China under 61861136003, 61621091 and 61673237, Beijing National Research Center for Information Science and Technology under 20031887521, and research fund of Tsinghua University - Tencent Joint Laboratory for Internet Innovation Technology.

\bibliography{main.bbl}
\bibliographystyle{aaai}
\end{document}